# Deep Learning Based Classification of Unsegmented Phonocardiogram Spectrograms Leveraging Transfer Learning

Kaleem Nawaz Khan[1], Faiq Ahmad Khan[1], Anam Abid[1,2], Tamer Olmez[3], Zumray Dokur[3], Amith Khandakar[4], Muhammad E. H. Chowdhury[4], Muhammad Salman Khan[1,5*]

[1] *AI in Healthcare, Intelligent Information Processing Lab, National Center of Artificial Intelligence, UET Peshawar, Pakistan;*

[2] *Department of Mechatronics Engineering, UET Peshawar, Pakistan;*

[3] *Department of Electronics and Communication Engineering, Istanbul Technical University, 34469, Istanbul, Turkey*

[4] *Department of Electrical Engineering, Qatar University, Doha-2713, Qatar;*

[5] *Department of Electrical Engineering (JC), University of Engineering and Technology, Peshawar, Pakistan;*

**Abstract**

Cardiovascular diseases (CVDs) are the main cause of deaths all over the world. Heart murmurs are the most common abnormalities detected during the auscultation process. The two widely used publicly available phonocardiogram (PCG) datasets are from the PhysioNet/CinC (2016) and PASCAL (2011) challenges. The datasets are significantly different in terms of the tools used for data acquisition, clinical protocols, digital storages and signal qualities, making it challenging to process and analyze. In this work, we have used short-time Fourier transform (STFT) based spectrograms to learn the representative patterns of the normal and abnormal PCG signals. Spectrograms generated from both the datasets are utilized to perform three different studies: (i) train, validate and test different variants of convolutional neural network (CNN) models with PhysioNet dataset, (ii) train, validate and test the best performing CNN structure on combined PhysioNet-PASCAL dataset and (iii) finally, transfer learning technique is employed to train the best performing pre-trained network from the first study with PASCAL dataset. We propose a novel, less complex and relatively light custom CNN model for the classification of PhysioNet, combined and PASCAL datasets. The first study achieves an accuracy, sensitivity, specificity, precision and F1 score of 95.4%, 96.3%, 92.4%, 97.6% and 96.98% respectively while the second study shows accuracy, sensitivity, specificity, precision and F1 score of 94.2%, 95.5%, 90.3%, 96.8% and 96.1% respectively. Finally, the third study shows a precision of 98.29% on the noisy PASCAL dataset with transfer learning approach. All the three proposed approaches outperform most of the recent competing studies by achieving comparatively high classification accuracy and precision, which make them suitable for screening CVDs using PCG signals.

* *Correspondence:* Muhammad Salman Khan; salmankhan@uetpeshawar.edu.pk

## 1. Introduction

Cardiovascular diseases (CVDs) are considered one of the major causes of death all over the world. Almost 75% of the developing countries are affected by CVDs [1]. In the United States nearly 34.3% and in Europe 48% of the deaths are related to CVDs [2, 3]. To detect CVDs, auscultation is considered as one of the most frequently used techniques. Auscultation is the process of listening the heart sound using a stethoscope. Physicians and cardiologists need extensive training to develop expertise in understanding the auscultation [4]. Moreover, it is important to note that the diagnosis accuracy of medical students and primary care physicians is between 20-40%, while the diagnosis accuracy of expert cardiologists is around 80% [5, 6].

Doctors and physicians employ computer aided diagnostic techniques for CVDs in which they consider patient's medical history, physical examinations and simple scoring technique. They interpret the results obtained from experiments based on their clinical experience. They use traditional taxonomy and tally each patient's data to the traditional trends and findings. This type of approach has proven inefficient, subjective and error-prone [7]. Moreover, all the emerging cardiovascular examination/treatment technologies are rapidly increasing their capacity and ability to capture huge amount of data and information. This rapid increase of technological complexity has made the work of cardiologists more demanding. This necessitates the development of automated screening techniques which provide cost-effective healthcare solutions without compromising the well-being of the patients.

There are two commonly-used publicly available datasets for automatic heart sounds classification problem, i.e., PhysioNet/CinC 2016 [8] and PASCAL challenge [9]. Consideration of these two datasets is due to the fact that they both have significant differences. Both have used different tools for data acquisition. They are recorded using different clinical protocols, and different overlapping noise that affects the signal qualities. In this paper, we have developed a novel, less complex and relatively light custom CNN model for PhysioNet dataset, which outperforms other competing models in terms of sensitivity, specificity and accuracy. The proposed model requires lesser number of convolutional layers, filters and fully connected layers to achieve same or better performance utilizing less computational resources and time. Subsequently, a deep learning based solution that leverages transfer learning concept is proposed for heart sound screening utilizing both the publicly available datasets. Phonocardiogram (PCG) signals produced by heart are fragmented into a fixed window size. Afterwards, the chunked signals are converted into time-frequency domain spectrogram which are classified using the two-dimensional (2D) CNN models.

The PASCAL dataset consists of limited number of PCG signals. Therefore, it is often recommended to use pre-trained deep learning models with such small dataset. It is due to the fact that deep learning networks tend to overfit with small number of training examples. The PASCAL dataset is low amplitude, noisy and very difficult to classify. Even the expert radiologists and doctors find it very difficult to annotate the dataset. The major part of the dataset, i.e. Dataset B is recorded using digital stethoscope. The highest accuracy with Dataset B is reported to be 59.78% using spectrogram images and pre-trained ResNet [10]. On the other hand, PhysioNet consists of large number of recordings for training which is suitable for deep learning models. Therefore, it is proposed to train a CNN model with PhysioNet dataset rather than using PASCAL dataset for training the CNN model. Another possibility is to combine both the datasets to a train a robust CNN model, which can be tested on both of these datasets and other potential datasets. Therefore, this study is designed to train a CNN model with the PhysioNet dataset which can be used to train, validate and test PASCAL dataset using transfer learning concept to improve the performance on the PASCAL dataset. In summary, the following experiments are performed to analyse the classification performance of the PCG spectrogram:

1) Train, validate and test different variants of convolutional neural networks with PhysioNet dataset to identify the best performing model structure

2) Train, validate and test the best performing CNN model with combined PhysioNet-PASCAL dataset

3) Utilize transfer learning technique to fine-tune the best performing PhysioNet-trained model and train, validate and test on PASCAL dataset

The manuscript has been divided into six sections. Section I introduces the problem statement while Section II introduces different machine learning and deep learning models and their application in the classification of PCG signals. Section III discusses about the datasets used in this study. Section IV presents the methodology, including the pre-processing techniques, dataset preparation, spectrogram generation and different proposed studies. Results of different studies and comparative performance with literature are discussed in Section V. The last section (Section VI) concludes the paper by presenting the usefulness of the study and possible future extensions.

## 2. BACKGROUND AND RELATED WORKS

Automatic PCG signal analysis has been widely focused in the recent years, specifically for automated segmentation and classification of heart sounds. It ranges from screening to diagnosis in the regions particularly suffering from the lack of healthcare services. Medical experts and researchers have studied practical and effective murmur classifier systems to improve the physicians' diagnostic accuracy. Moreover, with the technological advancement, machine learning algorithms for automating detection of heart auscultation have emerged. Also, mobile phone applications and digital stethoscopes come into play to record heart sounds. Machine learning models have been extensively used to classify the PCG signals [11]. The automatic, effective and accurate heart murmur classification techniques are required to identify heart related disorders. The following subsections discuss classical as well as deep convolutional neural network-based solutions for automated PCG signal classification.

### *2.1 CLASSICAL MACHINE LEARNING ALGORITHMS*

Machine learning (ML) is a widely used approach for automated heart murmur classification and for the identification of heart related abnormalities. Most of the recent studies have used multi-layer perceptron (MLP), support vector machines (SVMs) and other classical machine learning algorithms for preprocessing [12-15]. All such solutions have used different types of preprocessing techniques to segment and extract the best PCG signal segments from the recording [16-18].

These types of solutions are not considered ideal as they require human intervention and they are not fully automated. Studies like [3, 15-18] have used MLP and SVM to classify heart murmurs, which are trained only on simulated heart sounds but demonstrate low accuracy, when tested with real heart sounds. Other research attempts have used simulated heart sound with small number of real heart sounds [33, 35, 36] which cannot be not considered a true representative of the real data.

Initially, conventional machine learning algorithms such as K-nearest neighborhood (KNN), MLP, SVM and random forest (RF) were used in the classification process of heart sounds [37]. For conventional machine learning algorithms one of the important steps was to extract features from the PCG signals and subsequently these features were fed to the classifiers. Later, with the powerful and more robust ML algorithms such as Convolutional Neural Network (CNN), the performance is further improved. CNN is employed as automatic feature extractor utilizing large amount of data [38, 39]. Specifically, with image datasets, CNN effectively applies convolution operations to calculate the features, and with the pooling operation contained in the CNN, down samples the image using adjacent pixel information. According to a study conducted by Liu et al. [28], deep neural networks were applied for the first time for automatic analysis of heart sounds in 2016 on PhysioNet Computing in Cardiology Challenge dataset. Recently, many attempts are made using deep neural networks for the segmentation and classification of heart sounds, as discussed in the following subsections.

Table 1 : Summary of the recent articles for the classification of PCG signal using classical machine learning models and convolutional neural networks.

| **Reference** | **Preprocessing/Features** | **Classifier** | **Results (Accuracy/ Precision or sensitivity and specificity)** | **Datasets** |
|---|---|---|---|---|
| Potes et al. [12] | 124 time-frequency features | AdaBoost and CNN | Sensitivity:94.24%<br>Specificity: 77.81% | 2016 PhysioNet/CinC |
| Ryu et al. [19] | Heart sound recordings are filtered using windowed-sinc Hamming algorithm, scaling and segmentation | DNN | Sensitivity: 70.8%<br>Specificity: 88.2% | 2016 PhysioNet/CinC |
| Kucharski et al. [20] | Time-frequency parameters | DNN | Sensitivity: 99%<br>Specificity: 91.6% | 2016 PhysioNet/CinC |
| Chen et al. [21] | Mel-frequency cepstral coefficients (MFCCs) | DNN | Accuracy: 91% | Private Dataset |
| Rubin et al. [22] | Time-frequency heat map representations | DNN | Specificity: 95%<br>Sensitivity: 73% | 2016 PhysioNet/CinC |
| Dominguiz et al. [23] | Neuromorphic auditory sensor for real time decomposition of audio into frequency bands | DNN | Accuracy: 94%<br>Specificity: 95.12%<br>Sensitivity: 93.20% | 2016 PhysioNet/CinC |
| Alaskar et al. [24] | Deep features | Pretrained AlexNet and SVM | Accuracy: 85% | 2016 PhysioNet/CinC |
| Malik et al. [25] | Extraction of fundamental heart sound cycles (FHSC) | DNN and SVM | Accuracy: 95%<br>Sensitivity: 98% | PASCAL |
| Siddique Latif et al. [26] | MFCCs | DNN | Specificity: 98.36%<br>Sensitivity: 98.86% | 2016 PhysioNet/CinC |
| Abduh et al. [27] | MFCCs | DNN | Accuracy: 95%<br>Specificity: 97%<br>Sensitivity: 89% | 2016 PhysioNet/CinC |
| Li et al. [28] | DWT, entropy and frectal | DNN | Specificity: 92%<br>Sensitivity: 96% | 2016 PhysioNet/CinC |
| Fatih Demir et al. [6] | Spectrogram generation, deep feature extraction | DNN + SVM | Precision for Dataset A 3.19,<br>Precision 0.79 for Dataset B | PASCAL |
| Wei Han et al. [29] | MFCC map of a segment | SNMFNet Classifier | Specificity: 77.66%<br>Sensitivity: 89.18% | 2016 PhysioNet/CinC |
| Krzysztof Wolk et al. [10] | Spectrogram | ResNet pre-trained network | 93% precision | PASCAL |
| Madhusudhan Mishra et al. [30] | 8 different statistical features | Stacked Autoencoder, CNN | Accuracy: 95%<br>Specificity: 94%<br>Sensitivity: 97% | Private Dataset |
| Miguel Sotaquira et al. [31] | 107 time-frequency domain features | DNN | Accuracy: 92.6%<br>Sensitivity: 91.3%<br>Specificity: 93.8% | 2016 PhysioNet/CinC |
| Ali Raza et al. [32] | 50,000 frame of each signal | RNN | Accuracy: 80.80% | PASCAL |
| Sinam Ajitkumar et al. [33] | Continuous wavelet transforms | DNN | Accuracy: 90%<br>Sensitivity: 90%<br>Specificity: 90% | 2016 PhysioNet/CinC |
| Tianya Li et al. [33] | Frequency features | DNN | Precision: 96% (Dataset A) | PASCAL |
| Bozkurt et al. [34] | MFCC and Mel-Spectrogram | DNN | Accuracy: 81.5% | 2016 PhysioNet/CinC |

## 2.2 *DEEP NEURAL NETWORKS*

Artificial Neural Network (ANN) is a mathematical model that mimics the structure of biological neural network (NN). The Artificial Intelligence (AI) community has proposed a diverse spectrum of NN models to solve the real-life emerging problems. An artificial neural network which consists of multiple layers, named as deep neural network (DNN) has become state-of-the-art tool and has shown outstanding results in wide range of application areas such as, speech processing and automatic speech recognition [40-42], computer vision [43-45]and ANN based PV power prediction [46]. The working principle of DNN involves using a large number of hidden layers where the output of the current layer is fed as an input to the next layer, which consequently strengthens the classification or regression capabilities of the network. DNN achieves great flexibility to learn and express the concepts/patterns in a hierarchical form [47]. The abstract concepts will be expressed in relation to the relatively less abstract. Like many other fields, DNN models have shown very prominent results in the field of medical image processing [45, 48, 49]. It is considered a useful approach to transform the heart sound into an image and DNNs can be used to classify the images. There are many recent studies, listed in TABLE 1, where this notion is applied and it has shown promising results. All these studies presented in TABLE 1 have used either their own designed dataset, or publically available datasets such as PASCAL or PhysioNet datasets. A DNN based approach is presented by Chen et al [21], in which Mel-frequency cepstral coefficients (MFCC) features are used for the classification. They have shown that as compared to KNN, ANN, Gaussian mixture models, logistic differentiation and SVM, the DNN models have achieved better classification results.

Convolutional Neural Network (ConvNet/CNN) is a Deep machine learning model which utilizes convolutional layers along with fully-connected layers and is being popularly used to classify 1D signals and 2D images. The main benefit of ConvNet is that it requires much less pre-processing compared to other machine learning paradigm. Moreover, the laborious effort of identifying correct features for better classification is no more required rather network has the ability to automatically learn those or more useful characteristics directly from the raw data. It has been reported in numerous literatures that a CNN model can reliably identify the spatial and temporal dependencies of an image with the application of relevant filters and the network can be better trained to understand the sophistication of the image provided that a large dataset is used for training.

CNN consists of one or more convolutional layers, max pooling layers and generally followed by one or a number of dense layers. The convolutional layers are for the problem specific and representative features extractor. The final layers which are often fully connected layers are employed for the classification purposes based on the features extracted by the low-level layers of the architecture. The most significant feature of CNN model is the capability of feature extraction. Unlike other classifiers and learning algorithms, it does not essentially require features to be fed for classification tasks. It possesses the complex features identification ability which are nearly impossible to design manually [50]. This aspect enables CNN to handle huge amount of high dimensional data that is exploited for large amount of representative features. The proposed approach has employed CNN model development from scratch and also deploys transfer learning aspect of CNN for the PCG signals classification.

Rubin et al. [22] have studied an algorithm for the automated classification of heart sounds, which combines the CNN and time-frequency spectrogram of PCG signals. Studies like [6, 10, 23] have employed the concept of spectrogram images. They have used PhysioNet/CinC Challenge 2016 dataset to train the CNN architecture and then performance of the respective methods was assessed on the test set provided by the challenge. All the three methods have accuracies of 94%, 93% and 75% respectively. Potes et al. in [12] have utilized 124 time-frequency features to train CNN and AdaBoost classification algorithms and have achieved 94.24% of sensitivity and 77.81% of specificity. Ryu et al. in [19] have utilized Hamming-algorithm-based filtered, scaled and windowed PCGs for the classification and have achieved 70.8% and 88.2% of sensitivity and specificity respectively, using CNN. Similarly, Kucharski et al. in [20] also have used CNN for the classification of PCG and time-frequency parameters as input to the CNN.

Besides time-frequency features, MFCCs are also widely used features for PCG analysis with the combination of CNN and other machine learning classification algorithms [51]. In [52], frequency features are captured using fast Fourier transform (FFT) and a convolutional based model is used to develop a CVD detection system. The experiments are performed with PASCAL dataset. Similarly, in [53] spectrograms are generated for PCGs and used as a representative feature set for the heart sound classification. The experiments are carried out with privately developed dataset. Employing Mel-spectrogram features, the authors in [54] have designed a reliable system for heart diseases diagnosis using PhysioNet dataset. The overall score of this study is reported as 81%.

Miguel Sotaquir et al. in [31] also used 107 time-frequency features as input to DNN and in addition, decision factor was incorporated to minimize the trade-off between sensitivity and specificity and higher scores were achieved. Jinghui Li et al. in [28] have proposed a wavelet fractal and twin support vector machine (TWSVM) algorithm for the feature extraction and classification. Different features are extracted and utilized (i.e. wavelets, energy entropy and fractal) and accuracy, sensitivity, specificity and F1 Score of 90.4%, 94.6%, 85.5% and 95.2% respectively, are achieved. Malik et al. in [25] have proposed an algorithm for automatic localization of fundamental heart sound and its classification using variable length window and peak amplitude threshold variation.

Apart from conventional feature extraction methods, many authors have used deep features using deep neural network for the PCG analysis [6, 24]. Wei Han et al. in [29] have used MFCC segment maps as features to develop Semi-Non-Negative Matrix Factorization Network (SNMFNet) classifier. They have considered low dimensional features from semi-non-negative matrix factorization (SNMF). Krzysztof et al. [10] and Akshaya Balamurugan et al. [55] have used spectrograms of heart sounds to train ResNet pre-trained network and Stacked Residual Net-works respectively. Both have achieved 63.04% and 90.4% of accuracy respectively. Stacked Autoencoder and CNN are used with eight statistical features as input to the networks by Madhusudhan Mishra et al. [30] and they have achieved specificity, sensitivity and accuracy of 94%, 97% and 95%. In [32], the authors have extracted 50,000 frames from each heart sound recording to have fixed sizes of 782 frames using decimate technique. Recurrent neural network (RNN) is applied as classifier and an accuracy of 80.80% was achieved. Sinam Ajitkumar et al. [33] have used continuous wavelet transform based scalogram as input to the pre-trained AlexNet CNN model and achieved an accuracy, sensitivity and specificity of 90%, 90% and 90%. CNN is one of the most successful models of DNN [56].

All the discussed techniques have contributed towards the automation and accurate detection of heart sound abnormalities. The proposed study is concerned with the CNN based classification of heart sounds focused on the spectrogram representation of PCG signals.

## 3. DATASET DESCRIPTION

PASCAL [9] and PhysioNet/challenge 2016 [8] are the two publicly available datasets for the PCG classification. These datasets are widely used in the literature to design automatic PCG abnormality detection systems. PhysioNet and PASCAL datasets contain 3,240 and 683 heart sound recordings respectively. Table 2 summarizes the details of these datasets. The PhysioNet dataset consists of 2,575 normal and 665 abnormal PCG signals, while the PASCAL dataset consists of 381 normal and 302 abnormal PCG signals, respectively. Two datasets are significantly different from each other. The heart sounds are recorded using different sensors, age group, ethnicity and recording environments. The PhysioNet dataset is imbalanced in terms of normal and abnormal recording numbers as mentioned above. The PhysioNet dataset has normal to abnormal ratio of 4:1 while that is for PASCAL dataset is almost 1:1. A comparison of the sample signals in both the time domain and time-frequency domain (using spectrogram) are shown in Figure 1. Both the datasets were recorded with different sampling frequency and therefore, the signals are down-sampled to 2000 Hz for a fair comparison. It can be seen from both the time and time-frequency domain plots that the PCG signals from PASCAL dataset are nosier than the PhysioNet data. Two major heart sounds, S1 and S2 are apparent in the normal PhysioNet recordings and the signal bandwidth ranges from 15 to 150 Hz. On the other hand, normal PASCAL recordings have heart sounds other than S1 and S2 and the signal bandwidth ranges from 15 to 500 Hz. This indicates that PASCAL dataset has noise superimposed on the PCG signals which can significantly affect the signal quality and making it harder to process, analyze and diagnose. Each dataset is further explained in the following subsections:

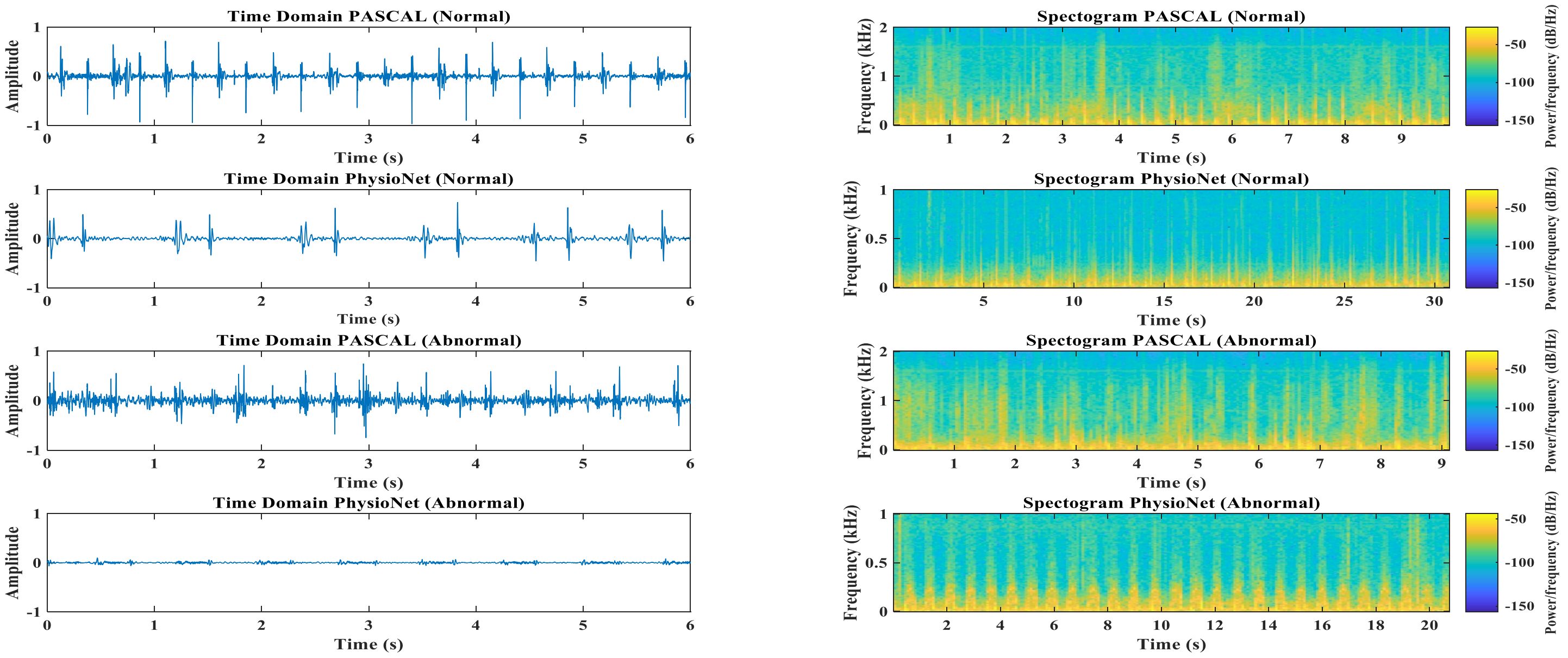

Figure 1. Sample Time domain (left) and Spectrogram (right) representation of the PCG signals from PASCAL and PhysioNet datasets.

### 3.1 PHYSIONET/CINC (2016)

This dataset was made available for public in 2016 for the classification of phonocardiograms challenge. The heart sounds were recorded at eight different sources by seven different research groups.

Overall, 4,430 sounds were recorded from 1,072 subjects. 3153 recordings from 764 subjects were made available for training and the remaining 1277 from 308 subjects were kept hidden as test set. The data was classified in three categories i.e., normal, abnormal and uncertain. There are signals ranging from 5 seconds to 120 seconds in length. The uncertain part of the dataset is often discarded and is not used for supervised classification. Since the signals are recorded in different setups and locations, significant noise is present in the recorded signals. The PCG signals have undergone pre-processing and filtration to remove the noise and non-usable segments in the recording. The final dataset used in the study is summarized in Table 2.

TABLE 2. STATISTICS OF THE PHYSIONET AND PASCAL DATASETS

| Database | | Total Recording | Normal | Abnormal |
|---|---|---|---|---|
| **PhysioNet Challenge 2016** | | 3,240 | 2,575 | 665 |
| **PASCAL** | Dataset A | 176 | 45 | 131 |
| | Dataset B | 507 | 336 | 171 |

*3.2 PASCAL*

PASCAL heart sound database was publicized in 2011 for the segmentation and classification challenges. It contains two subsets of heart sound recordings i.e., Dataset A and Dataset B. Auscultations in Dataset A are recorded using iStethoscope Pro iPhone app, whereas sounds in Dataset B are recorded using digital stethoscope, DigiScope. Dataset A contains four types of heart sounds: normal, murmurs, artifacts and extra heart sounds, while heart sounds in Dataset B are classified in three categories: Normal, murmurs and extra-systole. The final dataset used in the study is summarized in Table 2.

## 4. PROPOSED METHODOLOGY

The proposed methodology for this study is depicted in Figure 2. It consists of the following subtasks: The first phase is to collect the publicly available datasets and organize them in a compatible form for further processing. The next step is to perform necessary preprocessing and noise removal on the datasets. After preprocessing, the signals are chunked into fixed size window of 8 seconds. The segmented windows of the datasets are converted to spectrogram using Short-time Fourier transform (STFT). The proposed methodology is divided into three studies which will be discussed later. The generated spectrograms from PhysioNet dataset were used to train different CNN variants and identify the best performing CNN model. Secondly, the best performing CNN model structure is trained, validated and tested on combined PhysioNet and PASCAL dataset. Finally, the best performing PhysioNet model is trained, validated and tested using transfer learning concept on PASCAL dataset. The following subsections discuss each step, in details:

*4.1 PREPROCESSING THE PCG SIGNALS*

Since the datasets were recorded in different environments using different data acquisition tools, the sampling rates of the signals are not uniform. To ensure the same sampling rate for all the recordings while maintaining the Nyquist Criterion, all the recordings are resampled at 2000 Hz. Sampling rate was kept below 5000 Hz to avoid unnecessary high frequency noises which could be embedded with the desired signal [57]. Previous study [58] shows that fundamental heart sounds and murmurs lie in the frequency range of 20 to 400 Hz. In order to obtain the required frequency ranges and eliminate the unwanted frequencies or noise, 4th order Butterworth bandpass filter with cut-off frequencies of 20 to 400 Hz was used as shown in Figure 3, which has been found effective in biomedical signals processing especially in PCG signal analysis [59]. Figure 4 shows samples of the PCG time domain and spectrograms before and after the filtering with the Butterworth bandpass filter shown in Figure 3.

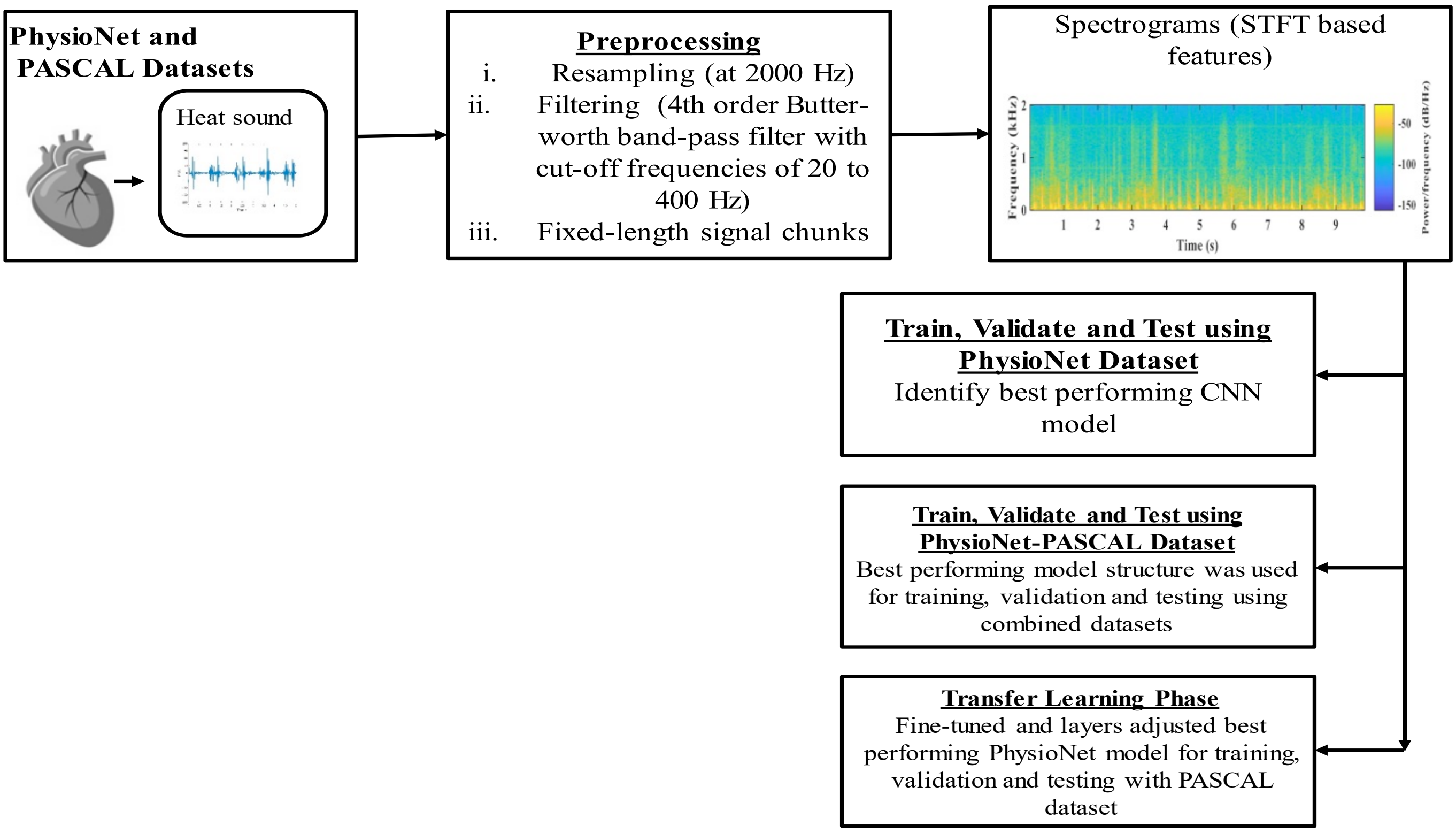

Figure 2. Methodology of the proposed work.

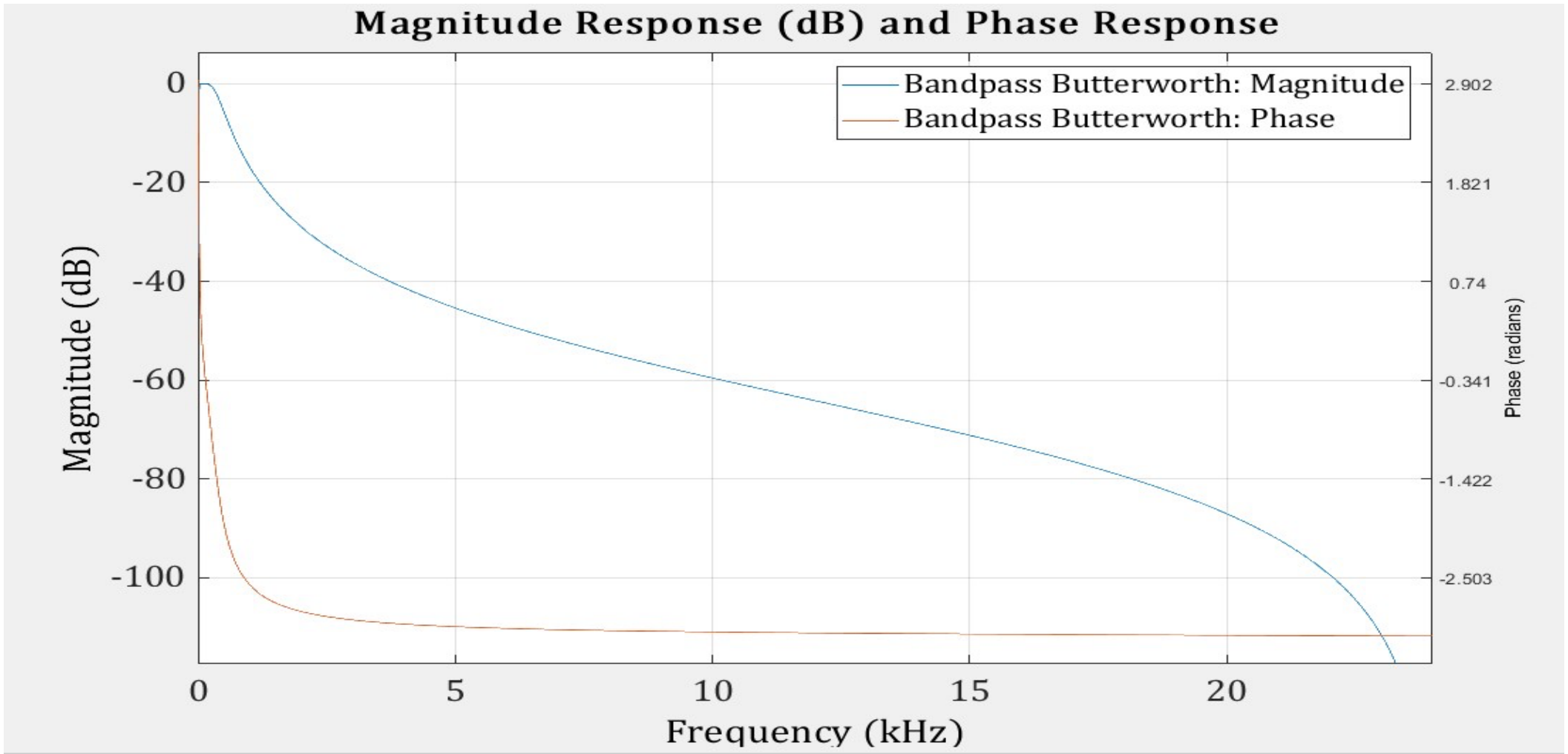

Figure 3. Magnitude and Phase Response of the Butterworth filter.

*4.2 DATASET PREPARATION*

The deep neural networks require larger set of training examples to be able to perform better. CNN requires fixed input size [60], and therefore, all the recordings are divided into 8 seconds segments (i.e., 16000 samples) as suggested by [51]. This approach not only fixes the variable length issue but also adds more training examples to the available datasets. The total number of PCG signals in PhysioNet dataset is increased from 3,240 signals to 7,221 segments. The two sub-sets of PASCAL datasets are merged into a single dataset. There are a number of PCG signals in the PASCAL dataset having length less than 8 seconds, which were discarded and long recordings were split into integer multiplication of 8 seconds segments. The fragmented recordings from PASCAL dataset were mainly from Dataset B (95% of the recordings) as most of the recordings in Dataset A are less than 8 seconds. After splitting into 8 seconds segments, the total number of recordings in the PASCAL dataset becomes 220. The third dataset which is named combined dataset is developed by combining both PhysioNet and PASCAL preprocessed datasets into a single set. The normal PCG signals are labeled as 0 while the abnormal signals are labeled as 1.

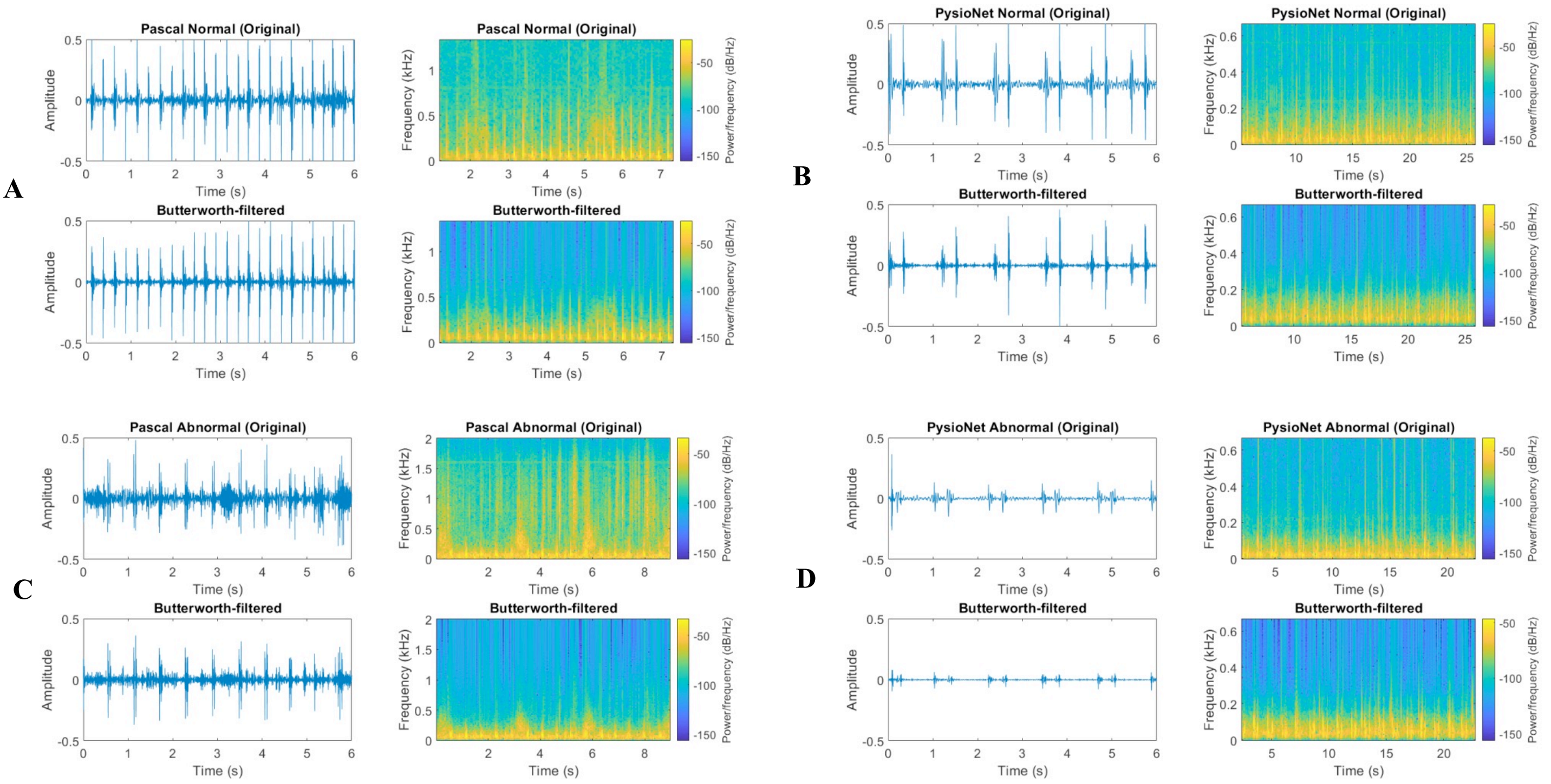

Figure 4. Time Domain and Spectrogram of A) Unfiltered and Filtered Normal PASCAL PCG, B) Unfiltered and Filtered Normal PhysioNet PCG, C) Unfiltered and Filtered Abnormal PASCAL PCG, D) Unfiltered and Filtered Abnormal PhysioNet PCG.

### *4.3 SHORT TIME FOURIER TRANSFORM ANALYSIS (SPECTROGRAM)*

Heart sounds are time-varying signals whose frequency properties change with respect to time [61]. A spectrogram shows how the frequency content of a signal changes over time and therefore each 8 second segments can be converted to a spectrogram image. This time-frequency domain analysis was performed to determine the dominant frequency components, which provide useful and distinctive insights in effective classification of the PCG signals. STFT is capable of capturing both temporal and spectral features of the signal by applying a small sized window to many small sections of the signal and calculate the Discrete Fourier Transform (DFT) instead of taking Fourier Transform of the whole signal at once [51]. STFT is mathematically represented as Equation 1 where x[k] denotes the original signal and h[k] represents the L-point window function.

$$X_{STFT}[m,n] = \sum_{k=0}^{L-1} x[k]h[k-m]e^{-\frac{j2\pi nk}{L}} \quad (1)$$

$$h(n) = \alpha + (1.0-\alpha)\cos\left[\frac{2\pi}{N}(n)\right] \quad (2)$$

$$H(\theta) = \alpha D(\omega) + \frac{1.0-\alpha}{2}\left[D\left(\omega - \frac{2\pi}{N}\right) + D\left(\omega + \frac{2\pi}{N}\right)\right] \quad (3)$$

$$H(n) = 0.54 - 0.46\cos(2\pi nN) \;\; 0 \le n \le N; N = L-1 \quad (4)$$

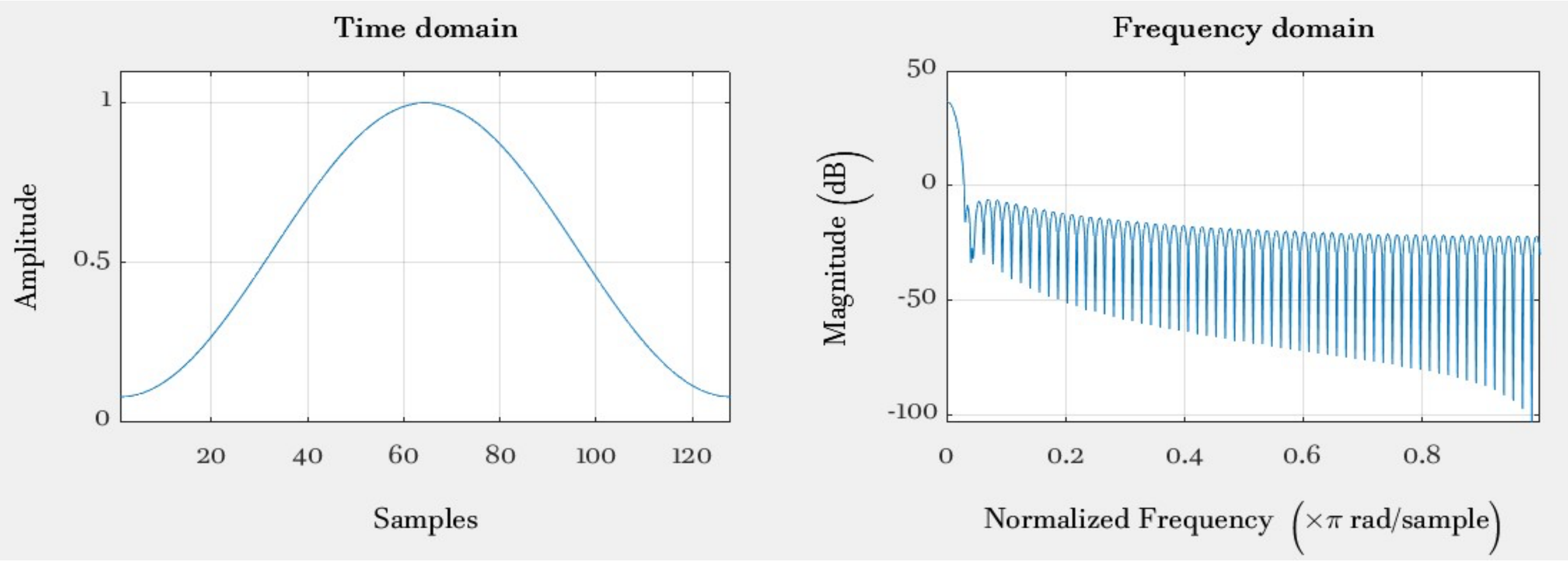

**Figure 5.** Hamming window in time and Frequency domain.

STFT based spectrograms are generated from the pre-processed and segmented signals. Each spectrogram is of the size of 137×310. The length for Fast-Fourier Transform was kept 128 to compute the spectrograms using a Hamming window of 64ms as shown in Figure 4, [51], [6]. The Hamming window is represented by Equation 2 with the corresponding spectrum is given by Equation 3. The parameter permits the optimization of the destructive side-lobe cancellation. In particular, when ◎ is adjusted to 25/46 (0.543478261...), the first side-lobe is canceled. The common approximation to this value of ◎ is 0.54, for which the window is called the Hamming window. The Hamming window is expressed in mathematical form in Equation 4.

### *4.4 EXPERIMENTAL PROCEDURE*

In this section, the details of the experimental procedures are presented. Three different studies were conducted in this work. All the datasets are randomly split into training, validation and test sets. The test set size is kept 10%, validation 15% and training 75% of the total size of the datasets. K-fold (with K=10) cross-validation is performed to make the results more robust. The input spectrogram image size of 137×310 is used for all the studies. The following section discusses each study individually:

TABLE 3. DETAILS OF THE DATASETS USED FOR TRAINING, VALIDATION AND TESTING.

| Studies | Types | No of Spectrogram/class | Training image /fold | Validation image /fold | Test image/ fold |
|---|---|---|---|---|---|
| Study 1 | Normal | 5501 | 4125 | 826 | 550 |
| | Abnormal | 1720 | 1290 | 258 | 172 |
| Study 2 | Normal | 5614 | 4211 | 842 | 561 |
| | Abnormal | 1827 | 1370 | 274 | 183 |
| Study 3 | Normal | 120 | 90 | 18 | 12 |
| | Abnormal | 100 | 75 | 15 | 10 |

TABLE 4. EXPERIMENTS WITH DIFFERENT ARCHITECTURAL VARIANTS OF CNN

| Experiment No. | Architectures of different investigated CNN models |
|---|---|
| 1 | Convolutional layers: 3 |
| | [(128x3x3)+maxpool(3x3)], [(256x3x3)+maxpool(3x3)], [(512x3x3)+maxpool(3x3)] |
| 2 | Convolutional layers: 3 |
| | [(128x3x3)+maxpool(3x3)], [(512x3x3)+maxpool(3x3)], [(128x3x3)+maxpool(3x3)] |
| 3 | Convolutional layers: 3 |
| | [(128x3x3)+maxpool(2x2)], [(256x3x3)+maxpool(2x2)], [(128x3x3)+maxpool(2x2)] |
| 4 | Convolutional layers: 4 |
| | [(128x3x3)+maxpool(2x2)], [(256x3x3)+maxpool(2x2)], (128x3x3), (64x3x3) |
| 5 | Convolutional layers: 5 |
| | [(96x11x11)+maxpool(3x3)], (256x5x5), (384x3x3), (384x3x3), [(256x3x3)+maxpool(3x3)] |
| 6 | Convolutional layers: 5 |
| | [(96x11x11)+maxpool(3x3)], [(256x5x5)+maxpool(3x3)], (384x3x3), (384x3x3), [(256x3x3)+maxpool(3x3)] |
| 7 | Convolutional layers: 6 |
| | [(16x3x3)+maxpool(3x3)], [(32x2x2)+maxpool(3x3)], (64x2x2), (128x2x2), [(256x2x2)+maxpool(3x3)], (256x2x2) |

## STUDY 1: BEST MODEL IDENTIFICATION USING PHYSIONET DATASET FROM DIFFERENT CNN VARIANTS

The first study is training, validation and testing of different architectural variants of CNN model, which are derived from the literature [12, 23, 25] using the PhysioNet dataset. Applying the search and tuning method specified by [62], the architectures are modified by tuning the number of filters, filter size and addition/removal of pooling layers. Table 3 shows the number of spectrogram images used for training, validation and testing of the different CNN models for each fold while Table 4 lists the different experimental models performed in this study. The details of the network architectures used by each experiment were listed. For all the experiments, the dataset of 7,221 images were used, where 5,501 and 1,720 images were normal and abnormal spectrograms respectively. After splitting into train and test sets, the available training data is 6,499 images which is further split into training and validation sets with 5,415 and 1,084 images respectively. After feeding all the training data to the networks, the training process continues for 110 epochs. The loss function and optimizer for these experiments are Binary cross-entropy and Adam respectively with learning rate of 0.001. The best performing model out of the seven different CNN variants was identified, which was used for the next two studies.

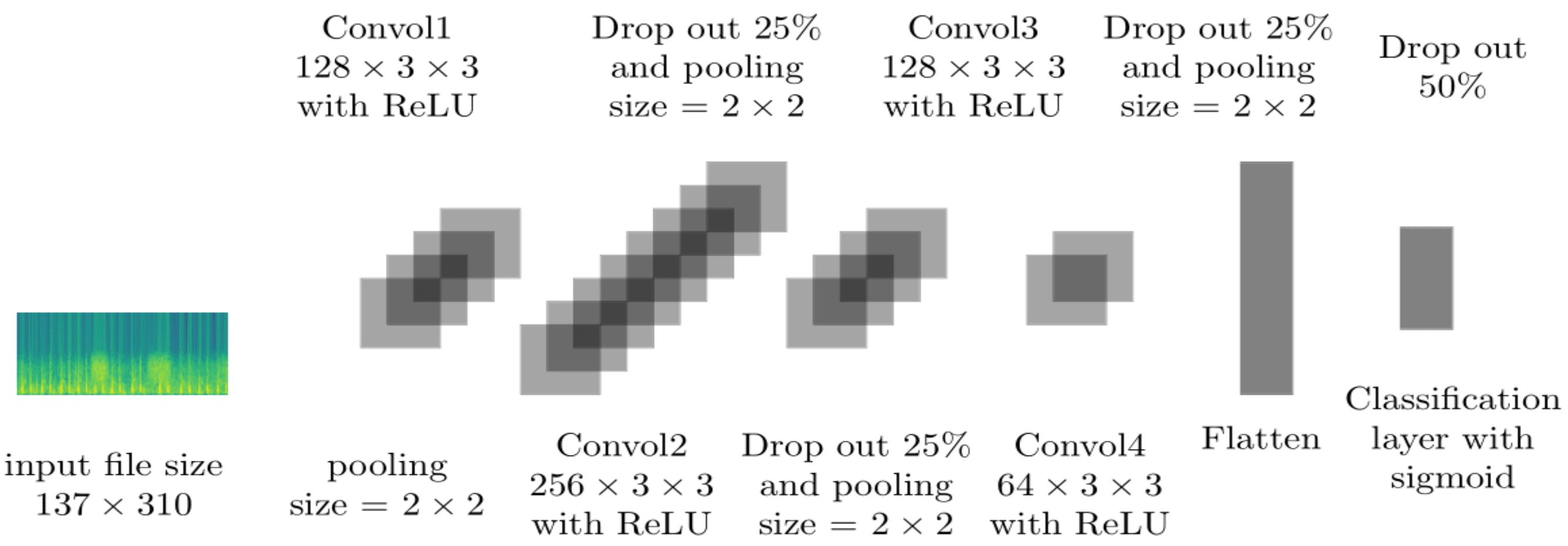

FIGURE 6. Proposed architecture of CNN model.

STUDY2: PERFORMANCE EVALUATION OF BEST MODEL STRUCTURE WITH COMBINED PHYSIONET-PASCAL DATASET

The second study was carried out with best performing CNN model structure from first study to train, validate and test PhysioNet-PASCAL merged dataset. The dataset consists of 7,441 (7,221 + 220) spectrograms in which 5,614 recordings are normal and 1,827 are abnormal. The total spectrogram available for training are 5,581 and 1116 images were used for validation while 744 images were used for testing. The network parameters and batch size were kept the same as the Study 1.

STUDY3: TRANSFER LEARNING USING PHYSIONET PRE-TRAINED MODEL AND EVALUATION ON PASCAL DATASET

Generally, for optimal training of CNN models require a larger dataset. However, the size of PASCAL dataset is limited. To address this issue, most of the studies in literature have used pre-trained network for PASCAL datasets. They have used pre-trained models such as AlexNet, VGG16 and ResNet15 trained on ImageNet dataset readily available in deep learning library of Python. Since these models are trained on images from the ImageNet dataset, they are not trained on spectrogram images. In contrary, in this study we have investigated the performance of the best model originally trained on PhysioNet dataset and trained, validated and tested on PASCAL dataset using transfer learning concept. The motivation of this is because both the PhysioNet and PASCAL datasets consist of heart sound recordings.

The PhysioNet model trained in Study 1 is tuned and modified to be used as a pre-trained network for PASCAL dataset. Total number of images used for training, validation and testing the pre-trained model were 165, 33 and 22 images respectively. Since the model was already pre-trained on spectrogram images and some of the layers are frozen, the risk of over-fitting was reduced.

## 5. Results and Discussions

In this section, the results from different studies will be reported along with the best performing model will be presented in details. Moreover, a comparative performance evaluation of different studies will be reported with the recent literatures. As mentioned earlier, Study 1, 2 and 3 were carried out on 722, 744 and 22 spectrogram images using PhysioNet, Combined PhysioNet-PASCAL and PASCAL only dataset respectively. The results reported in this section are the overall performance from K-fold cross-validation. Table 5 summarizes the training and test accuracies from 7 different models evaluated on PhysioNet dataset. Few of the experiments (experiment 1 & 2) show overfitting, where the training accuracies are above 90%, however the test accuracies are not acceptable. One of the reasons for overfitting is large number of filters in the middle layers, i.e., 512. Other experiments do not exhibit overfitting but their accuracy is comparatively less than the highest score achieved. The architecture with the highest accuracy was a comparatively less complex and light architecture as illustrated in experiment no. 4 (Table 4) and shown in Figure 6. It can be observed that removing almost half of the filters in the middle and adding maxpooling layers only to the starting layers make the proposed architecture a better option among all the experimented solutions.

The architecture of the proposed best performing CNN classifier for this article is shown in Figure 6. The network takes spectrogram as input in the input layer. The network consists of four convolutional layers with 128, 256, 128 and 64 filters respectively.
The training and validation accuracy plot for the best performing model on PhysioNet dataset is shown in Figure 7. Each layer is using filter size of 3x3 and Rectified Linear Unit (ReLU) activation function. Each convolutional layer is followed by a max pooling layer and then a dropout layer with 25% drop out. The maxpooling kernel size is 2x2 for all the layers. After the last dropout layer, the output is flattened in a single vector. The flattened output is fed to fully connected classification layer with 50% drop out. The classification layer is using sigmoid activation function as the sigmoid function comparatively performs well for binary classification problems [10].

TABLE 5. TRAINING AND TEST ACCURACY OF THE EXPERIMENTS WITH DIFFERENT ARCHITECTURAL VARIANTS OF CNN

| Experiment No. | | 1 | 2 | 3 | 4 | 5 | 6 | 7 |
|---|---|---|---|---|---|---|---|---|
| Results | training acc | 95.43% | 95.43% | 80.7% | 98% | 97.21% | 96.66% | 82% |
| (acc: accuracy) | test acc | 60.7% | 60.7% | 75.21% | 95.3% | 92.45% | 90.1% | 75% |
| | overfitting | Yes | Yes | No | No | No | No | No |

The best performing CNN structure were trained, validated and tested on combine dataset and training and validation accuracy plot for the model on combined dataset is shown in Figure 8. For PhysioNet dataset and combined PhysioNet-PASCAL dataset the highest accuracies using the best performing model were 95.3% and 94.3% respectively.

The results show that increasing PhysioNet dataset size by employing the strategy of fixed size windows, and fine tuning the model structure improves the detection performance. The same architecture which showed the best results for PhysioNet dataset, was failing initially for PASCAL dataset if the training was carried out from scratch, because the PASCAL dataset size is limited. Three different approaches are applied to train the proposed CNN architecture with the PASCAL dataset. First one is to train CNN solely with PASCAL datasets and then test the trained model on the test set. Since the amount of data is limited, these techniques did not perform well. To overcome this problem, two strategies are applied to improve the poor performance: (i) combining PhysioNet and PASCAL datasets for training, validation and testing, improves overall performance compared to training, validation and testing with PASCAL dataset alone; and (ii) the use of pre-trained model (trained with PhysioNet dataset), which is further fine-tuned and trained, validated and tested with PASCAL dataset. This approach significantly improves the results. Figure 9 shows the confusion matrix for Study 1-3 with the best performing models.

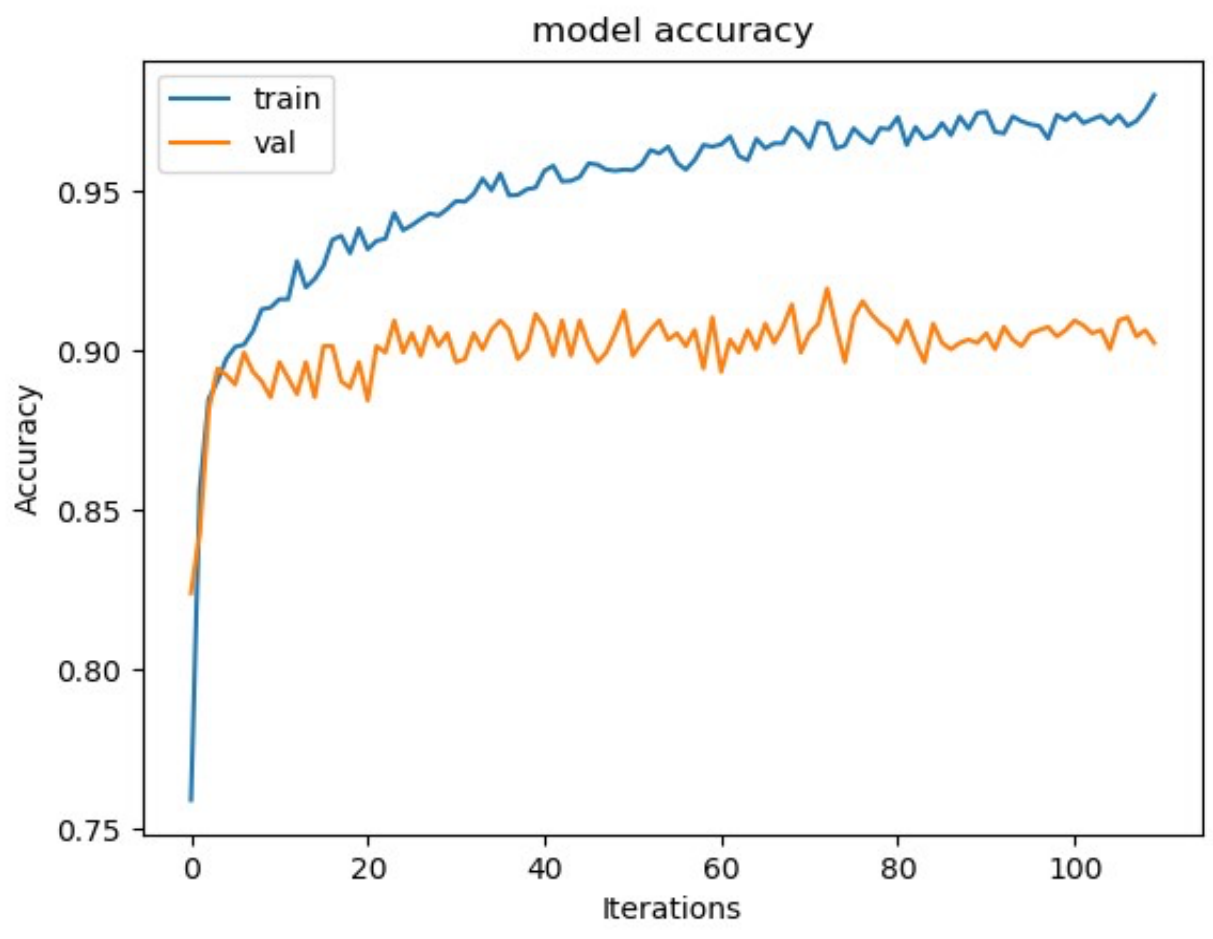

Figure 7. Training and validation with PhysioNet dataset

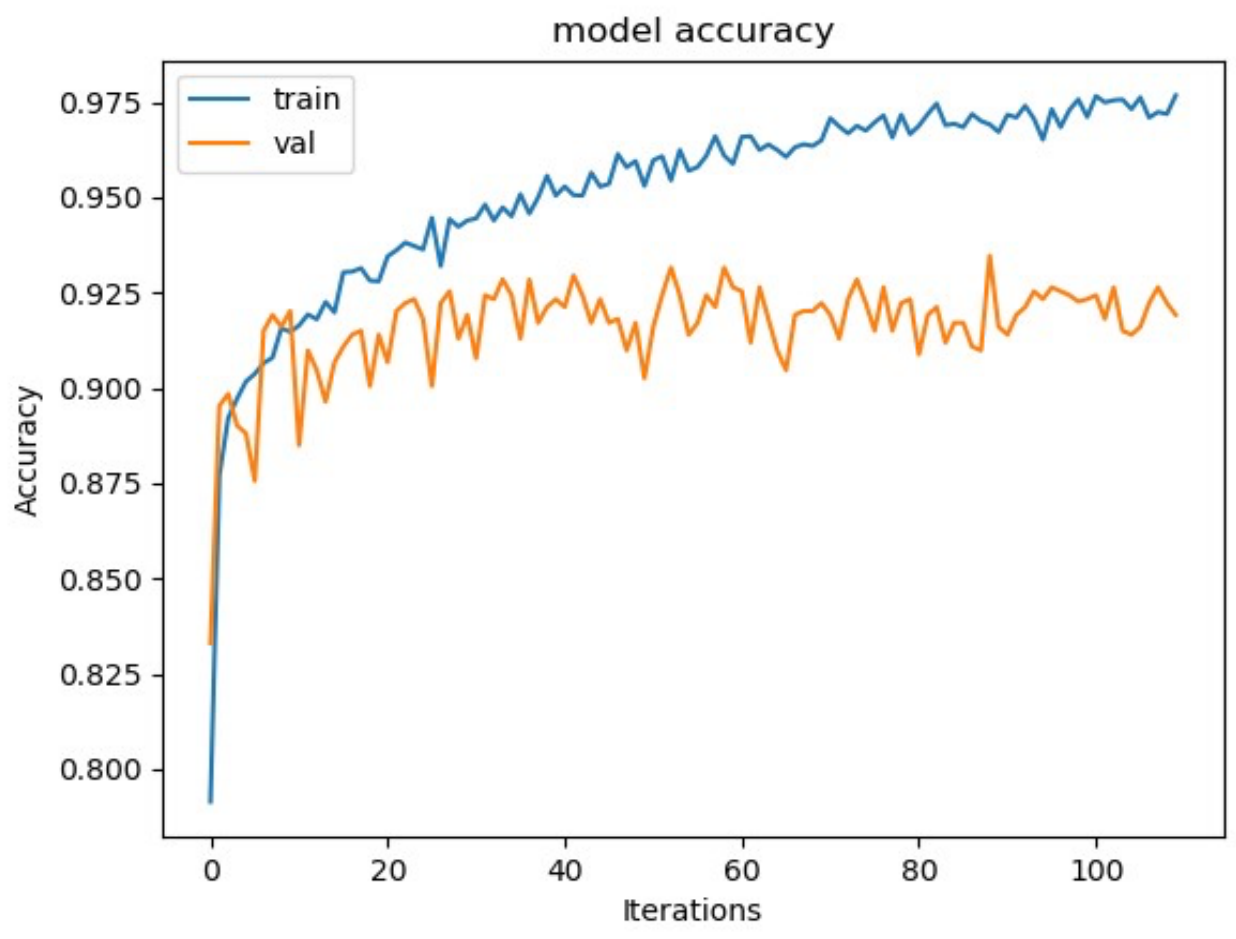

Figure 8. Training and validation with combined PhysioNet-PASCAL dataset.

TABLE 6. COMPARATIVE EVALUATION OF CNN MODELS WITH PHYSIONET DATASET

| Author (Year) | Accuracy | Sensitivity | Specificity | Precision | F1 Score |
|---|---|---|---|---|---|
| Potes et al. [12] (2016) | 86.02% | 94.24% | 77.81% | - | - |
| Rubin et al. [22] (2017) | 84% | 73% | 95% | - | - |
| Dominguiz et al. [23] (2017) | 97% | 93.20% | 95.12% | - | - |
| Miguel et al. [31] (2019) | 92.6% | 91.3% | 93.8% | - | - |
| Singh et al. [33] (2019) | 90% | 93% | 90% | - | - |
| Madhusudhan et al. [30] (2019) | 95% | 97% | 94% | - | - |
| Palani et al. [62] (2020) | 86% | 87% | 85% | - | - |
| **Proposed Approach** | **95.4%** | **96.3%** | **92.4%** | **97.6%** | **96.98%** |

## 5.1 COMPARISON WITH EXISTING CNN APPROACHES FOR PHYSIONET DATASET

The results produced by the proposed architecture with PhysioNet dataset are compared with some of the recent exiting studies using same dataset. Table 6 shows the comparative performance of the best performing CNN model on PhysioNet database with the recent literatures. Our proposed approach outperforms almost all the existing studies with the highest accuracy while consuming limited resources and preprocessing efforts. The existing studies either used dedicated hardware [25], pre-trained models [63] [64] or data is segmented for R-R intervals [21] before training. Our approach outshines with minimal involvement of such efforts. The sensitivity and specificity are slightly less than [65]. In [65], segmentation of R-R is involved, based on Shannon energy for the computation of S1 and S2. In contrast, the proposed approach is independent of the PCG segmentation. Unsegmented data is utilized rather only 8 second windows are generated, which accelerates the diagnosis process and simultaneously generates significant performance. Moreover, with combined PhysioNet and PASCAL dataset, the best performing model structure has achieved accuracy, sensitivity, specificity, precision and F1 score of 94.2%, 95.5%, 90.3%, 96.8% and 96.1% respectively.

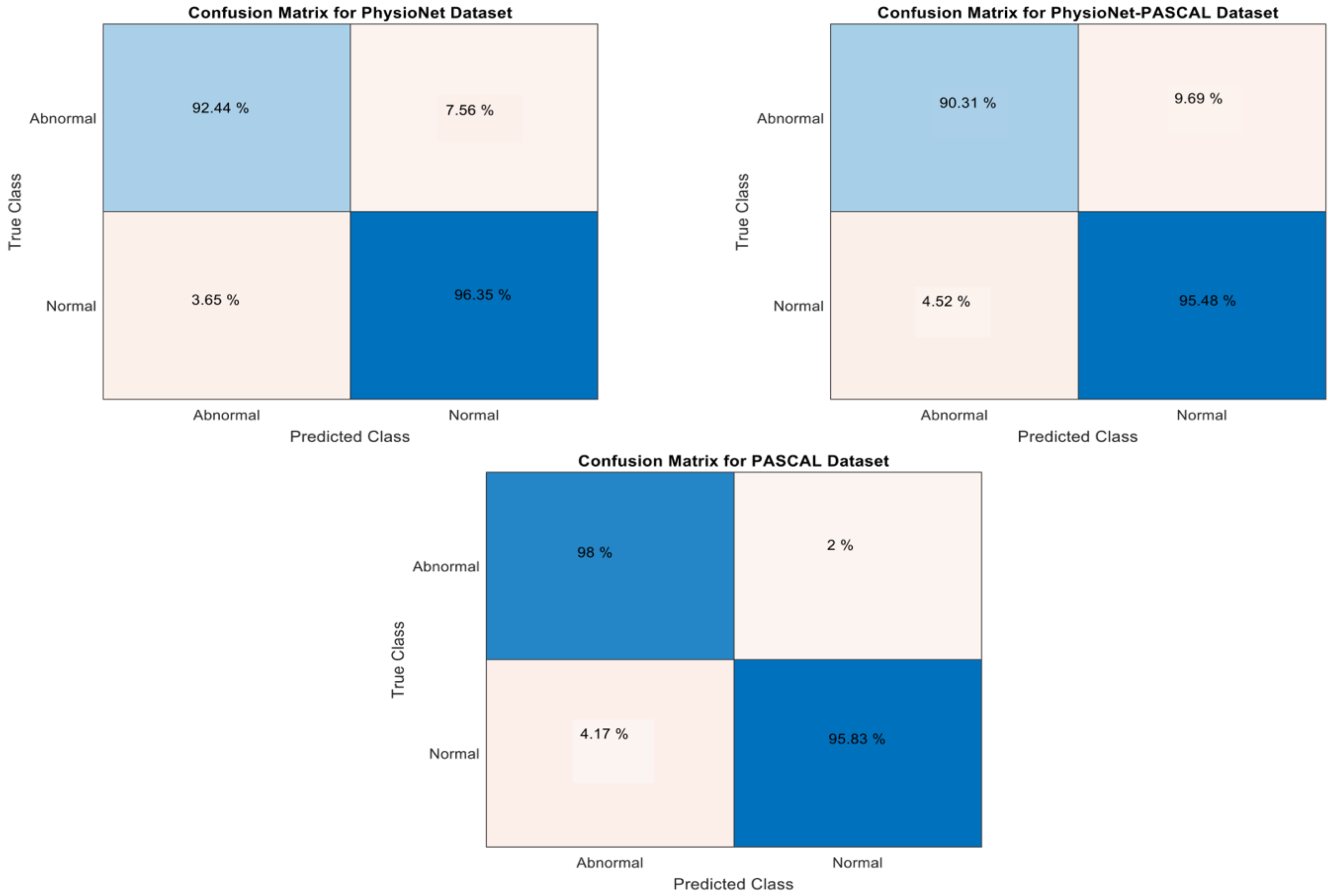

Figure 9. Confusion matrix for the best performing model.

## 5.2 COMPARISON WITH EXISTING CNN APPROACHES FOR PASCAL DATASET

PASCAL is one of the most complex and noisy datasets available for heart sounds classification and abnormality detection. As already discussed earlier, a portion of the dataset is recorded via phone while rest of the sounds are recorded using digital stethoscope. Existing studies have used different preprocessing and features extraction techniques to improve results with PASCAL datasets. The best performing CNN model pre-trained with PhysioNet dataset used to train, validate and test on PASCAL dataset. This transfer learning approach boosted the results significantly. It was observed that the accuracy, sensitivity, specificity, precisions and F1 score of spectrogram detection using this transfer learning approach were 96.8%, 95.8%, 98%, 98.29% and 97.05% respectively, which outperforms the best performing literatures with marginal difference. The approach of training CNN with PhysioNet dataset and then fine tune the model and using transfer learning technique improves model's performance significantly. It is evident from the results that the suggested strategy surely improves the results and outperforms other models which employ pre-trained models trained with datasets from different domains. However, it cannot be generalized to any arbitrary dataset as the PASCAL dataset is very small. In the future, this transfer learning concept can be evaluated an independent large dataset to evaluate the generalization capability of the pre-trained model.

TABLE 7: COMPARATIVE EVALUATION OF CNN MODELS WITH PASCAL DATASET

| Author (Year) | Accuracy | Sensitivity | Specificity | Precision | F1 Score |
|---|---|---|---|---|---|
| Deng et al. [66] (2016) | 76% | 34% | 95% | 76.5% | 74% |
| Zhang et al. [63] (2017) | 74% | 49% | 84% | 74% | 30% |
| Sujadevi VG et al. [64] (2017) | 74.4% | 69% | 87.85 | 69% | 77.7% |
| Wenjie Zhang et al. [67] (2017) | 71% | 51% | 80% | 71% | 67% |
| Fatima et al. [68] (2018) | 71.5% | 81% | 67% | 72% | 32% |
| Fatih Demir et al. [6] (2019) | 80% | 100% | 69% | 83 % | 33% |
| **Proposed Approach** | **96.8%** | **95.8%** | **98%** | **98.29%** | **97.05%** |

As discussed in Section III, the PASCAL dataset has got overlapped noise which significantly affects the signal quality and making it harder to process, analyze and diagnose. Therefore, the most of the existing techniques listed in Table 7 have less detection precision for abnormality as compared to normality detection. The approach used in [64], comparatively performs better in case of abnormality detection. Abnormality detection rate of the proposed approach is nearly the same as [64] whereas the normality detection rate of the proposed approach is 25% higher than the one reported in [64].

## 6. Conclusion and Future Work

In this study, different CNN architectures are utilized to perform extensive experimentation for preliminary screening and detection of heart rhythm abnormalities. The spectrogram representation of the heart sounds is utilized to provide input to the convolutional neural network architectures to learn representative features. These spectrograms are generated using STFT analysis of PCG signals from an 8 second window. Different experiments were performed with different variations and modifications in the architecture and datasets. The experiments are carried out using the two widely used and publically available datasets, i.e., PASCAL and PhysioNet. The CNN architecture is designed and refined after several iterations of training and fine tuning. The designed model is also trained and validated with combined PhysioNet-PASCAL dataset to study the robustness of the proposed approach. The proposed best performing model with PhysioNet dataset achieves the highest score and outshines most of the existing approaches. Another contribution of this paper is the proposition of pre-trained network that is trained with PhysioNet network and tuned for PASCAL dataset. In the existing literature the conventional pre-trained models are utilized for PASCAL dataset. We believe this is the first attempt that pre-trained network with PhysioNet is used for PASCAL dataset. The setup achieves the best results in terms of accuracy and precision and outperforms the existing CNN based studies for PASCAL dataset.
The possible extension of this study is to utilize the same concept for the R-R interval segmented heart sounds. This will be useful to provide CNN more precise and accurate representation to learn useful patterns and insights, which will potentially boost the results and performance of the proposed approach further. Another possible extension is to utilize the time domain representation and features with the same architecture setup to focus on both time and frequency features to design broader analysis of PCG signals for CVDs screening.

## Funding

The Higher Education Commission (HEC) of Pakistan through Artificial Intelligence in Healthcare, Intelligent Information Processing Lab, National Center of Artificial Intelligence provided support for the work and the claims made herein are solely the responsibility of the authors.

## Acknowledgments

Authors acknowledge the support of NVIDIA Corporation with the donation of the Titan X Pascal used for this research.